\documentclass{article} 
\usepackage{iclr2025_conference,times}
\usepackage{graphicx}

\usepackage{amsmath,amsfonts,bm}

\def\eqref#1{equation~\ref{#1}}

\def\1{\bm{1}}

\DeclareMathAlphabet{\mathsfit}{\encodingdefault}{\sfdefault}{m}{sl}
\SetMathAlphabet{\mathsfit}{bold}{\encodingdefault}{\sfdefault}{bx}{n}

\usepackage{soul} 
\usepackage{booktabs}
\usepackage{hyperref}
\usepackage{url}
\usepackage{multirow}
\usepackage{graphicx}
\usepackage{caption}
\usepackage{authblk}
\usepackage{xcolor}
\usepackage[table]{xcolor}
\usepackage{algorithm}
\usepackage{algorithmic}
\usepackage{amsmath,amssymb}
\usepackage{textcomp}
\usepackage{marvosym}
\setcitestyle{numbers, square}  

\makeatletter
\renewcommand\NAT@open{[}
\renewcommand\NAT@close{]}
\makeatother

\definecolor{InkBlue}{RGB}{25,25,112}   
\definecolor{Burgundy}{RGB}{128,0,32}  
\definecolor{ForestGreen}{RGB}{0,85,45}

\newcommand{\name}{\textsc{Faithful Contouring}}

\title{\name: Near-Lossless 3D Voxel Representation Free from Iso-surface }

\iclrfinalcopy 

\author{
        Yihao Luo$^{1\dag *}$,
        Xianglong He$^{2 \dag}$,
        Chuanyu Pan$^{3}$,
        Yiwen Chen$^{3,4}$,
        Jiaqi Wu$^{5}$,
        Yangguang Li$^{6}$,
        Wanli Ouyang$^{6}$,
        Yuanming Hu$^{3}$,
        Guang Yang$^{1\ddag}$,
        ChoonHwai Yap$^{1 \ddag*}$
    \\
    {\normalsize $^{1}$Imperial College London}~~
    {\normalsize $^{2}$Tsinghua University}~~
    {\normalsize $^{3}$Meshy}~~
    {\normalsize $^{4}$Nanyang Technological University}~~\\
    {\normalsize $^{5}$Mathmagic}~~
    {\normalsize $^{6}$The Chinese University of Hong Kong}
    \\
    \tt \color{red}{\href{https://github.com/Luo-Yihao/FaithC}{https://github.com/Luo-Yihao/FaithC}}
}

\begin{document}

\maketitle
\begingroup
\addtocounter{footnote}{-2}
\renewcommand\thefootnote{}
\footnotetext{
\begin{minipage}{\linewidth}
\scriptsize
\hspace{-6mm}
\dag~Equal contribution ~ \ddag~Co-last authors ~ *~Corresponding authors ~\href{mailto:y.luo23@imperial.ac.uk}{\Letter~y.luo23@imperial.ac.uk}%
\end{minipage}
}
\endgroup

\vspace{-8mm}
\begin{figure*}[h]
  \centering
  \includegraphics[width=\linewidth]{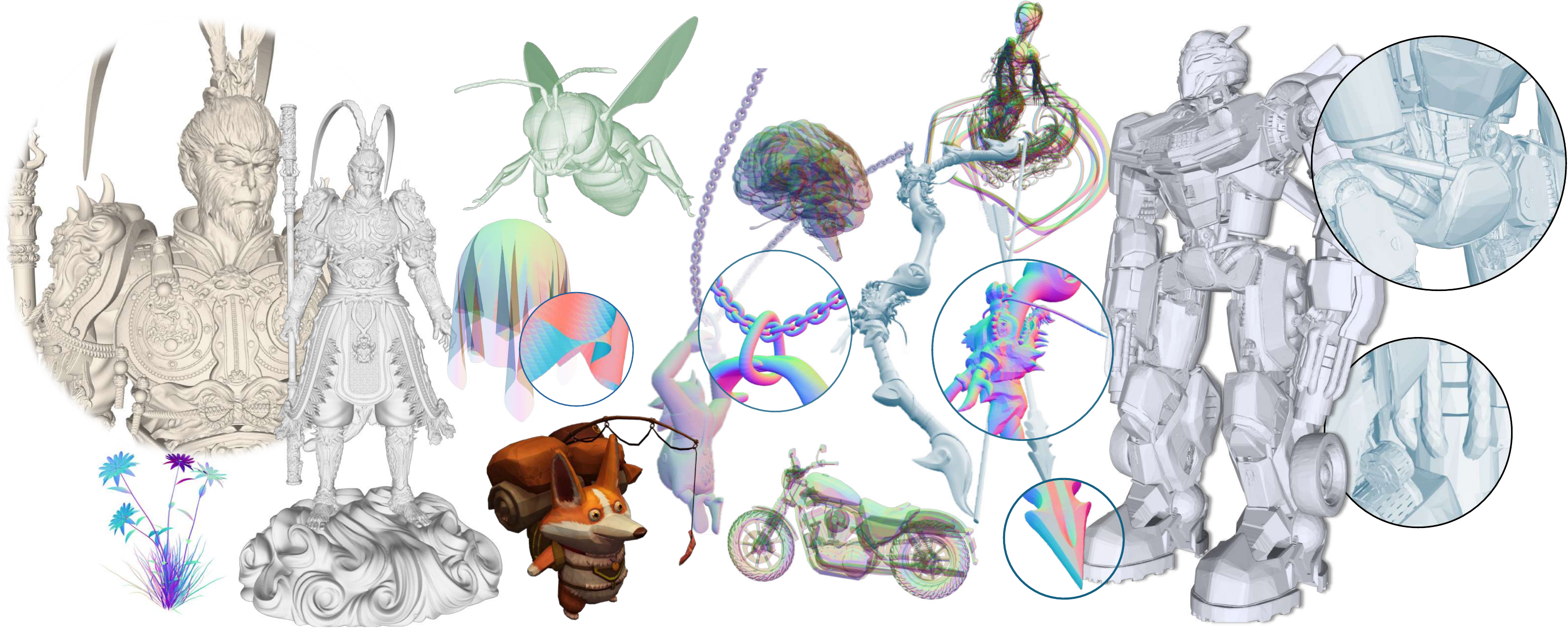}
  \caption{
    \textit{\textbf{\name}: A Near-Lossless Voxelized 3D Representation keeps fine-grained geometric details while maintaining internal structure. This representation encodes an arbitrary mesh into voxelized tokens, supporting 2048+ resolution with neither iso-surface extraction from the converted SDFs nor differentiable rendering optimization.  
    Please zoom in to view the detailed geometry from the remeshing results.}
  }
  \label{fig:teaser}
\end{figure*}
\vspace{-2mm}

\begin{abstract}
\vspace{-2mm}
Accurate and efficient voxelized representations of 3D meshes are the foundation of 3D reconstruction and generation. However, existing representations based on iso-surface heavily rely on water-tightening or rendering optimization, which inevitably compromise geometric fidelity. We propose Faithful Contouring, a sparse voxelized representation that supports 2048+ resolutions for arbitrary meshes, requiring neither converting meshes to field functions nor extracting the isosurface during remeshing. It achieves near-lossless fidelity by preserving sharpness and internal structures, even for challenging cases with complex geometry and topology. The proposed method also shows flexibility for texturing, manipulation, and editing. Beyond representation, we design a dual-mode autoencoder for Faithful Contouring, enabling scalable and detail-preserving shape reconstruction. 
Extensive experiments show that Faithful Contouring surpasses existing methods in accuracy and efficiency for both representation and reconstruction. For direct representation, it achieves distance errors at the $10^{-5}$ level; for mesh reconstruction, it yields a 93\% reduction in Chamfer Distance and a 35\% improvement in F-score over strong baselines, confirming superior fidelity as a representation for 3D learning tasks.

\vspace{-3mm}
\end{abstract}

\section{Introduction}

High-fidelity 3D reconstruction and generation have become central problems in computer vision, graphics, and medical imaging, with wide-ranging applications in virtual/augmented reality~\cite{zhao2022metaverse}, robotics~\cite{xiang2017posecnn,wohlhart2015learning}, world models for environment understanding~\cite{ha2018world,gupta2023visual, engelcke2019genesis}, and embodied intelligence where agents interact with complex 3D environments~\cite{anderson2018vision,duan2022survey}. A fundamental prerequisite for these applications is an accurate voxelized representation of 3D shapes, which provides a regular discretization of space and facilitates efficient learning on tensor-based architectures~\cite{choy20163d,wu20153d,riegler2017octnet}. By normalizing irregular and nonlinear geometry into structured grids, voxel data enable scalable training of deep neural networks, in contrast to meshes or point clouds that require specialized operators, and remain one of the most robust foundations for 3D learning.

\textbf{Challenges of existing representations.}  
Distance-field representations such as occupancy and signed distance fields (SDFs)~\cite{park2019deepsdf,mescheder2019occupancy,chen2019learning} provide continuous functional descriptions of geometry, and neural implicit models like Occupancy Networks~\cite{mescheder2019occupancy} and NeRFs~\cite{mildenhall2020nerf} have extended this paradigm to learning-based reconstruction. Considerable effort has been devoted to improving surface quality, from early alternatives to marching cubes such as Dual Contouring~\cite{ju2002dual} to adaptive schemes like FlexiCubes~\cite{zheng2022flexicubes}. Nevertheless, all SDF- and occupancy-based approaches ultimately depend on watertight geometry and discretization-based remeshing, making them ill-defined for real-world meshes and fundamentally limiting fidelity~\cite{wang2021neus,boulch2022poco}. Moreover, distance fields are intrinsically global and nonlinear: evaluating the sign of a point requires global operations such as winding number~\cite{barill2018fast,jacobson2013robust} or flood-fill consistency, which are computationally expensive and error-prone for open or non-manifold surfaces. As a result, sharp features and internal structures are frequently lost, and generated meshes are often restricted in quality and usability.

Another major line of work formulates 3D reconstruction through differentiable image formation, most prominently via volumetric ray marching or rasterization surrogates~\cite{kato2018neural,niemeyer2020differentiable,liu2020neural}. While effective for supervision from 2D images, these methods are inherently constrained by discretization resolution and field-of-view: limited sampling along rays leads to blurred or aliased geometry, and reconstruction quality degrades with insufficient multi-view coverage~\cite{yariv2020multiview,martin2021nerf,verbin2022refnerf}. Because geometry remains implicit, extracting a usable surface still requires remeshing through marching cubes or related algorithms~\cite{yariv2021volume,wang2021neus, atanasov2022neural}, which is difficult to scale and typically yields meshes of limited fidelity.

Beyond these specific drawbacks, both distance-field and rendering-based representations share a more general limitation: 1. They do not naturally support structural manipulation and editing, making it difficult to enable tasks such as selective filtering, splitting, or compositional assembly in downstream 3D learning.

\vspace{-2mm}
\begin{figure}[!h]
  \centering
   \includegraphics[width=\linewidth]{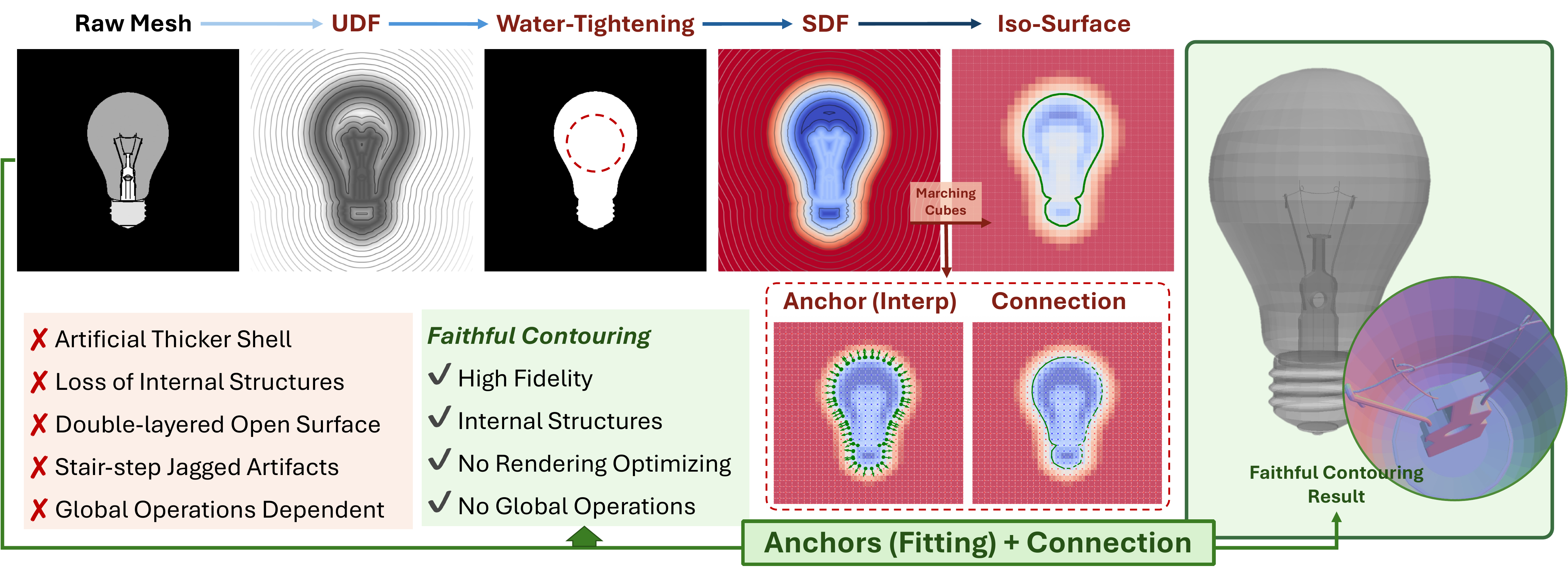}
   \vspace{-5mm}
\caption{
\textbf{Comparison of representing pipelines.} \textit{Traditional \textbf{UDF $\rightarrow$ water-tightening $\rightarrow$ SDF $\rightarrow$ iso-surface} pipelines, relying on Marching Cubes and its variants, introduce artifacts at each lossy step, including artificial surface thickening, loss of internal structures, and jagged iso-surface extraction. In contrast, \name{} directly obtains voxelized features, including fitted anchors and connections, from raw meshes with a highly accurate remeshing algorithm.}}

\label{fig:contouring_introduction}
\end{figure}
\vspace{-2mm}

\textbf{What future 3D generative models need.}  
Recent advances in generative modeling have significantly expanded the frontier of 3D content creation. 
Sparse-voxel and structured latent methods such as 3DShape2VecSet~\cite{zhang20233dshape2vecset}, Michelangelo~\cite{micheangelo2024},Clay~\cite{zhang2024clay}, Trellis~\cite{xiang2025structured}, Hi3DGen~\cite{ye2025hi3dgen}, Sparc3D~\cite{sparc3d2025}, SparseFlex (TripoSF)~\cite{he2025triposf}, and Ultra3D~\cite{ultra3d2025} demonstrate the ability to generate high-resolution geometry with fine details and arbitrary topology. 
Parallel efforts in large-scale diffusion backbones, including 
LRM~\cite{hong2023lrm}, MeshLRM~\cite{wei2024meshlrm}, InstantMesh~\cite{xu2024instantmesh}, and Hunyuan3D~\cite{zhao2025hunyuan3d}, highlight the feasibility of scaling text- or image-conditioned 3D generation to full textured assets. 

Together, these works illustrate a clear trend toward more expressive and scalable 3D generative frameworks grounded in voxelized representations, but they also reveal shared representational bottlenecks. \textcolor{Burgundy}{\textit{Regardless of how advanced the network architectures become, explicit or implicit SDF representations followed by Marching Cubes–style remeshing remain the de facto path to mesh reconstruction}}. This pipeline introduces geometric inaccuracies, restricts the resolution under $1024$, and struggles to capture complex internal structures without incurring rendering losses.

These limitations motivate the need for a voxelized representation obtained \textbf{directly} from arbitrary raw meshes—rather than through a converted distance field—that can \textbf{losslessly} preserve smoothness, sharpness, and internal details, while maintaining voxel regularity to support structural operations in deep learning tasks. These limitations motivate a voxelized representation derived directly from raw meshes.

\textbf{Our contributions.}  
Motivated by these challenges, we present \textit{\textbf{Faithful Contouring}}, an \textbf{almost-lossless}, \textbf{distance-field-free} voxelized representation that directly encodes meshes into sparse voxelized features  \textbf{without rendering optimizations} and \textbf{global operations} (therefore, GPU parallel computation friendly). Our contributions are:
\begin{itemize}
    \item A high-fidelity voxelized 3D representation, supporting \textbf{2048+} resolutions, capable of robustly handling \textbf{open surfaces}, \textbf{non-manifold elements}, \textbf{multi-component assemblies}, and \textbf{complex topologies}, paired with an \textbf{efficient remeshing algorithm} preserving \textbf{sharp edges} and \textbf{internal structures}.
    \item Retains the \textbf{standardization of voxels} while enabling  \textbf{texturing, manipulation, and editing}, such as affine transformations, filtering, and assembly, ensuring high flexibility for downstream applications.
    \item A dual-mode reconstruction architecture integrating sparse 3D convolutions with lightweight attention, supporting either faithful auto-compression from voxelized representation into itself or convenient conversion from point-cloud input.
    \item Experiments show that our method achieves \textbf{state-of-the-art performance} both at the representation level — preserving sharp geometry and internal structures of raw data — and at the reconstruction level, surpassing existing methods in both accuracy and efficiency.  
\end{itemize}

\section{Related Work}

For decades, a large body of work has represented 3D shapes using various signed distance field (SDF)–based designs, either explicitly or implicitly. However, these approaches almost invariably rely on Marching Cubes (MC)~\cite{lorensen1987marchingcubes} to extract mesh surfaces, which often introduces ambiguity and smoothing artifacts. Variants of MC such as Dual Contouring~\cite{ju2002dual}, FlexiCubes~\cite{zheng2022flexicubes}, and sparse voxel schemes~\cite{ren2024xcube, sparc3d2025} attempt to improve sharpness and efficiency, yet all distance-field methods follow a similar pipeline: a raw mesh is first converted into a watertight proxy, then assigned inside/outside labels to form an SDF, and finally remeshed via MC or its variants.  

Each stage in this pipeline introduces loss, largely because it relies on global heuristics rather than local voxel evidence. 
\emph{Watertight preprocessing} often employs $\epsilon$-ball dilation to seal gaps, which can undesirably alter topology. 
\emph{Sign assignment} infers inside/outside for non-watertight meshes using global methods such as flood-fill, winding numbers~\cite{jacobson2013robust}, or rasterization statistics~\cite{barill2018fast}; these techniques are sensitive to mesh degeneracies and unstable near non-manifold or open regions, and they inevitably discard internal cavities. 
Because such procedures cannot be resolved by parallel local computation, they impose fundamental limitations on scalability, resolution, and efficiency. 
Finally, \emph{surface extraction and remeshing} (e.g., MC with subsequent regularization) further over-smooths the geometry, attenuates high-frequency details, and often produces voxel-lattice artifacts.

Implicit representations including Occupancy Networks~\cite{mescheder2019occupancy}, DeepSDF~\cite{park2019deepsdf}, and NeRF~\cite{mildenhall2020nerf} define geometry continuously and can be trained from sparse data. 
Differentiable rendering extensions~\cite{kato2018neural,niemeyer2020differentiable,liu2020neural} further allow learning from 2D images, with tri-plane features~\cite{chan2022eg3d} improving efficiency. 
Nevertheless, these methods remain bound by sampling resolution, field-of-view constraints, and implicit surface extraction, which hinder preservation of sharp details and internal structures.

Recent diffusion-based pipelines have achieved impressive single-image or sparse-view reconstruction, including Zero123++~\cite{liu2023zero}, One-2-3-45~\cite{liu2023one2345},
One-2-3-45++~\cite{liu2023one2345}, DMV3D~\cite{xu2023dmv3d}, LucidDreamer~\cite{liang2024luciddreamer}, Wonder3D~\cite{long2023wonder3d}, and GaussianDreamer~\cite{yi2024gaussiandreamer}. 
At a larger scale, foundation-style backbones such as LRM~\cite{hong2023lrm}, LGM~\cite{tang2024lgm}, Textto3d~\cite{chen2024text}, 3DGen~\cite{gupta20233dgen}, MeshLRM~\cite{wei2024meshlrm}, InstantMesh~\cite{xu2024instantmesh}, and Hunyuan3D~\cite{zhao2025hunyuan3d} extend controllable generation to high-resolution textured assets. 
In parallel, structured latent and sparse voxel approaches are rapidly emerging: 3DShape2VecSet~\cite{zhang20233dshape2vecset}, Clay~\cite{zhang2024clay}, Michelangelo~\cite{micheangelo2024}, Trellis~\cite{xiang2025structured}, Hi3DGen~\cite{ye2025hi3dgen}, SparseFlex (TripoSF)~\cite{he2025triposf}, Sparc3D~\cite{sparc3d2025}, and Ultra3D~\cite{ultra3d2025}, which highlight scalability and geometric fidelity while supporting part-level or multi-modal control. 
These works demonstrate the rapid progress toward expressive and scalable 3D generative frameworks, but remain fundamentally constrained by their underlying representations, limiting faithful preservation of sharpness, internal structures, and structural operability.

\section{Method: Faithful Contouring}
\subsection{FCT Representation and Remeshing}
Recall the two key steps of marching cubes and dual contouring~\cite{lorensen1987marchingcubes, ju2002dual}: (i) Interpolate the coordinates
of reconstructed vertices on the iso-surfaces from SDFs. (ii) Determine the connection to generate faces according to the sign changes, as shown in Fig.~\ref{fig:contouring_introduction}. 
A natural question arises: 

{\textit{Is there any way to directly extract candidate vertices (anchors) within each voxel from the raw mesh, and then reconstruct faces by determining the connection to achieve a marching-style remeshing? Rather than convert meshes into distance fields and then extract the iso-surface.}}

\begin{figure}[!ht]
  \centering
  \includegraphics[width=\linewidth]{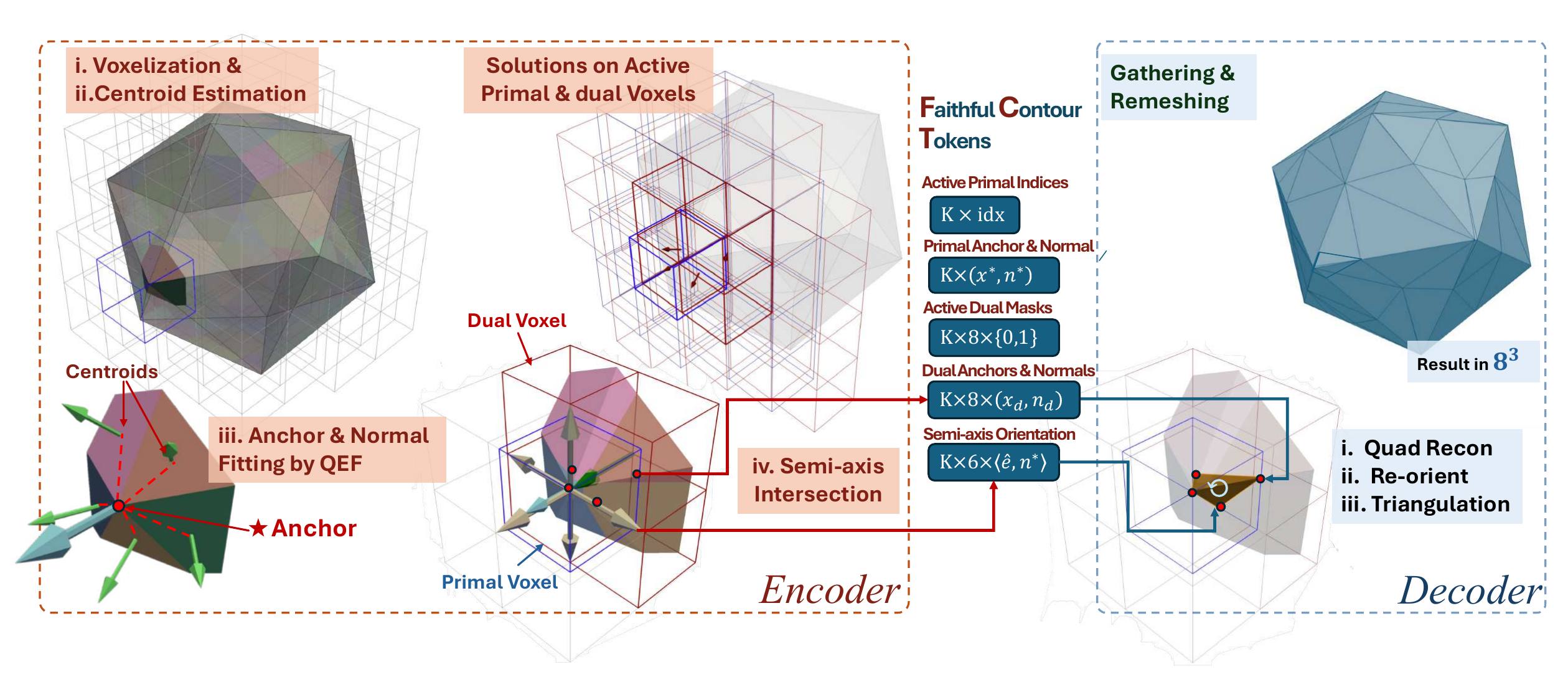}
  \vspace{-8mm}
  \caption{\textbf{Faithful Contour pipeline.} 
  \textit{Encoder voxelizes the input mesh, then computes centroids, anchors, and semi-axis intersections, and stores them in the Faithful Contour Token (FCT) on $K$ active voxels. 
  Decoder gathers anchors, resolves orientations, and remeshes the tokens into high-fidelity surfaces.  }}
  \label{fig:method}
\end{figure}

This is intuitively the idea behind \name{}, which converts a raw triangle mesh directly into a faithful sparse voxel representation and eliminates lossy intermediate to enhance fidelity.
The process is illustrated in Fig.~\ref{fig:method} and consists of two stages: (i) \emph{Encoder} solves the anchor position for each intersected voxel and records the connectivity by directed intersections of semi-axis to raw mesh; (ii)
\emph{Decoder} gathers anchors and assembles them as orientated faces conforming to  the connectivity information.
Unlike iso-surface extractions, the procedure is sign-distance-free and does not require a manifold hypophysis; therefore, it is naturally adaptive to arbitrary meshes, including open surfaces,  multi-components, and internal cavities.

\vspace{-4mm}
\vspace{-4mm}

\paragraph{Encoder (i): Active Voxel Detection.}
Let the input mesh be $\mathcal{M}=(\mathcal{V},\mathcal{F})$ where $\mathcal{V}$ represents vertices and $\mathcal{F}$ stands for triangle faces, and the voxel grid $\mathcal{G}$ consist of axis-aligned cubes. 
For each voxel $v\in\mathcal{G}$ and triangle $f\in\mathcal{F}$, overlap is tested by the Separating Axis Theorem (SAT)~\cite{akenine2001fast}. 
Projections on 13 axes (box axes, triangle normal, and edge–axis cross products) determine whether $f$ intersects $v$, marking $v$ as an \emph{active primal voxel}.

\paragraph{Encoder (ii): Intersection Centroids.}
For each active voxel–triangle pair, the clipped polygon $Q_{v,f}=v\cap f$ is obtained via sequential clipping against voxel planes~\cite{sutherland1974reentrant}. The centroid
\[
\mathbf{c}_{v,f} = \frac{1}{3A}\sum_{k=2}^{m-1} A_k (\mathbf{q}_1+\mathbf{q}_k+\mathbf{q}_{k+1}), 
\quad A_k=\tfrac{1}{2}\|(\mathbf{q}_k-\mathbf{q}_1)\times(\mathbf{q}_{k+1}-\mathbf{q}_1)\|
\]
is guaranteed to lie inside $v$ by the convexity of both voxels and triangle faces,
where $\{\mathbf{q}_1,\dots,\mathbf{q}_m\}$ are the ordered vertices of the polygon $Q_{v,f}$,
$m$ is the number of vertices,
$A_k$ is the area of the triangle $(\mathbf{q}_1,\mathbf{q}_k,\mathbf{q}_{k+1})$,
and $A=\sum_{k=2}^{m-1} A_k$ is the polygon area.
Each centroid is paired with the triangle normal $\mathbf{n}_f$, yielding reliable geometric samples $(\mathbf{c}_{v,f},\mathbf{n}_f)$.
\begin{algorithm}[!h]
  \caption{Faithful Contour Encoding}
  \label{alg:fct-encode}
  \begin{algorithmic}[1]
    \REQUIRE Mesh $\mathcal{M}=(\mathcal{V},\mathcal{F})$, voxel grid $\mathcal{G}$
    \ENSURE Faithful Contour Tokens (FCT)

    \FOR{each voxel $v\in\mathcal{G}$}
      \FOR{each triangle $f\in\mathcal{F}$}
        \IF{SAT detects overlap~\cite{akenine2001fast}}
          \STATE Clip $f$ to get polygon $Q_{v,f}$~\cite{sutherland1974reentrant}
          \STATE Compute centroid $\mathbf{c}_{v,f}$ and normal $\mathbf{n}_f$; add to sample set $\mathcal{S}_v$
        \ENDIF
      \ENDFOR

      \STATE Fit anchor position by solving $(M^\top M+\lambda I)\mathbf{x}^\ast=M^\top \mathbf{d}+\lambda \bar{\mathbf{c}}$ and $\bar{\mathbf{c}}=\tfrac{1}{N}\sum_i \mathbf{c}_i$ for both primal and dual voxels; Estimate the closed form average normals $\mathbf{n}^\ast$; Set mask $\mathbf{m}_d=1$ for each valid dual anchors else $\mathbf{m}_d=0$

      \FOR{each primal semix-axis $\hat{\mathbf{e}}$}
        \STATE Detect semi-axis crossings with Möller–Trumbore~\cite{moller1997fast} 
        \STATE Determine the direction by 
        $\mathrm{orient} = \mathrm{sign}\!\big(\langle \mathbf{n}^\ast,\hat{\mathbf{e}}\rangle\big)$
      \ENDFOR

    \STATE Append record $\big[v,(\mathbf{x}^\ast,\mathbf{n}^\ast),\{\mathbf{m}_d,(\mathbf{x}_d,\mathbf{n}_d)\}_{d=1}^8,\{\mathrm{orient}_e\}\big]$ to FCT
    \ENDFOR
    \STATE \textbf{return} (FCT)
  \end{algorithmic}
\end{algorithm}
\vspace{-2mm}

\begin{figure}[!ht]
  \centering
\includegraphics[width=0.8\linewidth]{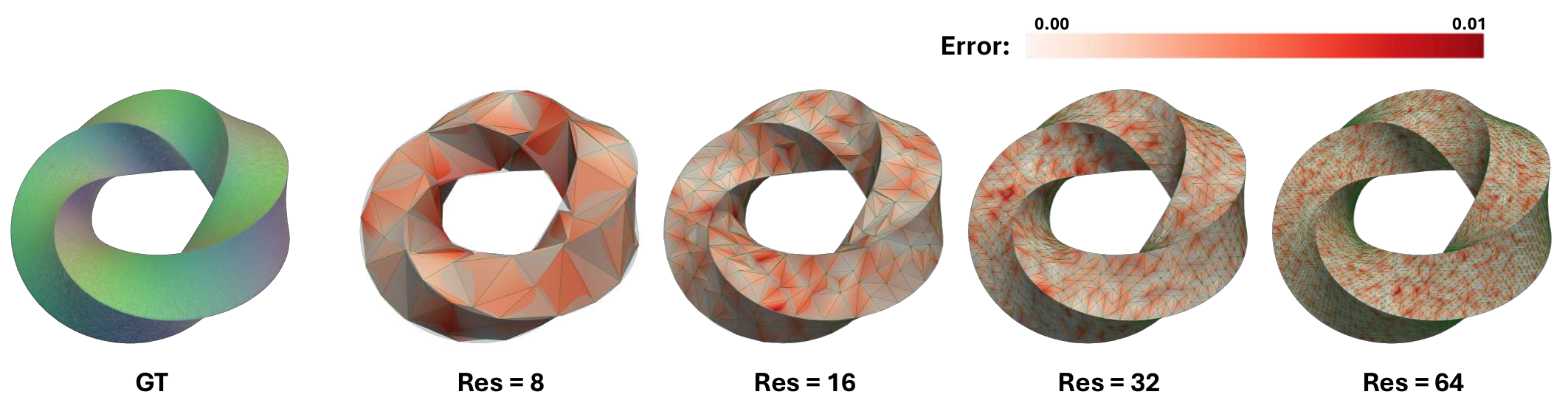}
  \vspace{-2mm}
  \caption{\textbf{Sharpness from low–resolution reconstruction with Faithful Contouring.} 
Ground-truth surface (GT) compared with reconstructions at voxel resolutions of $8^3$, $16^3$, $32^3$, and $64^3$. 
Despite coarse discretization, our method preserves overall shape and captures sharp geometric features, with error visualized in red.}
  \label{fig:lowpoly}
\end{figure}

\paragraph{Encoder (iii): Anchor Fitting.}
Each active voxel (or its eight duals) accumulates samples $\{(\mathbf{c}_i,\mathbf{n}_i)\}$. 
Anchor position and orientation are jointly estimated by quadratic error minimization:
\begin{align}
\mathbf{x}^\ast &= \arg\min_{\mathbf{x}}
\sum_i (\mathbf{n}_i^{\!\top}(\mathbf{x}-\mathbf{c}_i))^2 + \lambda \|\mathbf{x}-\bar{\mathbf{c}}\|^2 ,
\quad \bar{\mathbf{c}}=\tfrac{1}{N}\sum_i \mathbf{c}_i , \\
\mathbf{n}^\ast &= \arg\min_{\|\mathbf{n}\|=1}
\sum_i (\mathbf{n}^\top(\mathbf{x}^\ast-\mathbf{c}_i))^2 + \mu \|\mathbf{n}-\bar{\mathbf{n}}\|^2 ,
\quad \bar{\mathbf{n}}=\tfrac{1}{N}\sum_i \mathbf{n}_i .
\end{align}
The positional term enforces consistency with tangent-plane constraints, while the centroid regularizer stabilizes under ambiguity. 
The normal term aligns orientation with local offsets while regularized toward the average normal. 
Together, these objectives counteract ill-posed cases caused by nearly parallel normals and bias anchors toward \textbf{sharp edges} and \textbf{salient corners}, even in low voxel resolutions, shown in Fig.~\ref{fig:lowpoly}.

\vspace{-2mm}
\vspace{-2mm}

\paragraph{Matrix form and closed-form solvers.}
Let $M\in\mathbb{R}^{N\times 3}$ stack row vectors $\mathbf{n}_i^{\!\top}$ and let $\mathbf{d}\in\mathbb{R}^{N}$ collect $d_i=\mathbf{n}_i^{\!\top}\mathbf{c}_i$. 
Then the position objective is
\[
\min_{\mathbf{x}}\;\|M\mathbf{x}-\mathbf{d}\|_2^2+\lambda\|\mathbf{x}-\bar{\mathbf{c}}\|_2^2,
\]
with normal equations
\[
(M^{\!\top}M+\lambda I)\,\mathbf{x}^\ast=M^{\!\top}\mathbf{d}+\lambda\,\bar{\mathbf{c}},
\]
solved stably (e.g., by Cholesky~\cite{golub2013matrix}) since $M^{\!\top}M+\lambda I\succ 0$ for $\lambda>0$. 
(With weights $w_i>0$, use $W=\mathrm{diag}(w_i)$: $(M^{\!\top}WM+\lambda I)\mathbf{x}^\ast=M^{\!\top}W\mathbf{d}+\lambda\bar{\mathbf{c}}$.)

For the normal, define offsets $\mathbf{v}_i=\mathbf{x}^\ast-\mathbf{c}_i$ and $C=\sum_i \mathbf{v}_i\mathbf{v}_i^{\!\top}\in\mathbb{R}^{3\times 3}$. 
The objective reads
\[
\min_{\|\mathbf{n}\|=1}\;\mathbf{n}^{\!\top}C\,\mathbf{n}+\mu\|\mathbf{n}-\bar{\mathbf{n}}\|_2^2
\;=\;\min_{\|\mathbf{n}\|=1}\;\mathbf{n}^{\!\top}(C+\mu I)\mathbf{n}-2\mu\,\mathbf{n}^{\!\top}\bar{\mathbf{n}}+\mu\|\bar{\mathbf{n}}\|_2^2.
\]
Setting the Lagrangian gradient to zero yields
\[
(C+\mu I)\,\mathbf{n}=\mu\,\bar{\mathbf{n}}+\lambda_{\!n}\,\mathbf{n}.
\]
A practical closed form is the Tikhonov-regularized~\cite{tikhonov1977solutions} solution followed by normalization:
\[
\tilde{\mathbf{n}}=(C+\mu I)^{-1}(\mu\,\bar{\mathbf{n}}),\qquad
\mathbf{n}^\ast=\frac{\tilde{\mathbf{n}}}{\|\tilde{\mathbf{n}}\|_2},
\]
(or, with weights $w_i$: $C=\sum_i w_i\,\mathbf{v}_i\mathbf{v}_i^{\!\top}$, $\bar{\mathbf{n}}=\tfrac{\sum_i w_i\mathbf{n}_i}{\sum_i w_i}$).
This yields a unique, well-conditioned normal even when $\{\mathbf{v}_i\}$ are nearly co-planar~\cite{eckart1936approximation,taubin1995estimating}.

\paragraph{Encoder (iv): Semi-axis Intersections.}
To capture directed surface crossings, we apply the Möller–Trumbore segment–triangle test~\cite{moller1997fast} along voxel semi-axes $\hat{\mathbf{e}}\in\{\pm x,\pm y,\pm z\}$. 
Orientation is defined as
\[
\mathrm{orient} = \mathrm{sign}\!\langle \mathbf{n}^\ast,\hat{\mathbf{e}}\rangle \in \{-1,0,1\},
\]
where $0$ indicates no crossing or near-parallel alignment. 
Each voxel thus encodes a compact semi-axis code in $\{-1,0,1\}^6$.

\paragraph{Faithful Contour Tokens (FCT).}
All information is stored row-wise in the \emph{Faithful Contour Tokens}:
\[
{\rm FCT} =
\big[\;\text{voxel index},\;(\mathbf{x}^\ast,\mathbf{n}^\ast),\;
   \{\mathbf{m}_d,(\mathbf{x}_d,\mathbf{n}_d)\}_{d=1}^8,\;
   \{\mathrm{orient}_e\}_{e\in\{\pm x,\pm y,\pm z\}}\;\big],
\]
where $\mathbf{m}_d\in\{0,1\}$ is a binary mask indicating whether dual $d$ carries a valid anchor.  
These sparse tokens contain all the information we need to reconstruct meshes and are presented in regular voxelized forms suitable for deep learning. Alg. \ref{alg:fct-encode} outlines the entire pipeline for obtaining FCT.

\begin{algorithm}[!h]
  \caption{Faithful Contour Decoding}
  \label{alg:fct-decode}
  \begin{algorithmic}[1]
    \REQUIRE Faithful Contour Tokens FCT
    \ENSURE Reconstructed mesh $\mathcal{M}'=(V',F')$

    \STATE \textbf{Global gather:} For each dual voxel $d$, average anchors across adjacent primals
    \STATE Construct vertex set $V'=\{\mathbf{x}_d\}$ with unified anchors
    \FOR{each primal face in FCT}
      \STATE Connected four incident dual anchors $\{\mathbf{x}_{d_1},\mathbf{x}_{d_2},\mathbf{x}_{d_3},\mathbf{x}_{d_4}\}$
      \STATE Re-orient quadrilateral using semi-axis code; reverse order if inconsistent
      \STATE Select diagonal minimizing facet normal deviation
      \STATE Add two divided triangles to face set $F'$
    \ENDFOR
    \STATE \textbf{return} mesh $\mathcal{M}'=(V',F')$
  \end{algorithmic}
\end{algorithm}
\vspace{-2mm}

\paragraph{Decoder: Gathering \& Remeshing.}
Decoding begins with a global gather step in which dual voxels shared by multiple primals are unified into single anchors.  
For each dual voxel $d$, the anchor position and normal are averaged over all incident primals:
\[
\mathbf{x}_d = \tfrac{1}{|\mathcal{P}(d)|}\sum_{p\in\mathcal{P}(d)} \mathbf{x}_d^{(p)}, 
\qquad
\mathbf{n}_d = \frac{\sum_{p\in\mathcal{P}(d)} \mathbf{n}_d^{(p)}}{\|\sum_{p\in\mathcal{P}(d)} \mathbf{n}_d^{(p)}\|},
\]
where $\mathcal{P}(d)$ denotes the set of primals adjacent to $d$.  
The resulting unified anchors collectively form the vertex set
\[
V' = \{\mathbf{x}_d \mid d \;\text{is a dual voxel with valid anchor}\},
\]
which directly serves as the reconstructed vertices. This averaging scheme provides a simple yet reliable strategy in practice; in principle, more refined aggregation rules, such as normal-weighted or area-weighted averaging, can also be employed.

On each primal face, four incident dual anchors 
$\{\mathbf{x}_{d_1},\mathbf{x}_{d_2},\mathbf{x}_{d_3},\mathbf{x}_{d_4}\}$ 
define a quadrilateral patch. The semi-axis code determines its orientation: if 
$\langle \mathbf{n}^\ast, \hat{\mathbf{e}}\rangle < 0$, anchor order is reversed.  
Triangulation is resolved by selecting the diagonal that minimizes normal deviation:
\[
\{d_i,d_j\} = \arg\min_{(1,3),(2,4)}
\sum_{t\in T_{ij}} \big(1 - \langle \mathbf{n}(t), \mathbf{n}_{\text{avg}} \rangle \big).
\]
All facets from quadrilaterals are assembled into the final mesh $\mathcal{M}'$. The pseudo-code for remeshing is presented as Alg. \ref{alg:fct-decode}.

\subsection{Editing of FCT}

\begin{figure}[!h]
  \centering
  \includegraphics[width=0.9\linewidth]{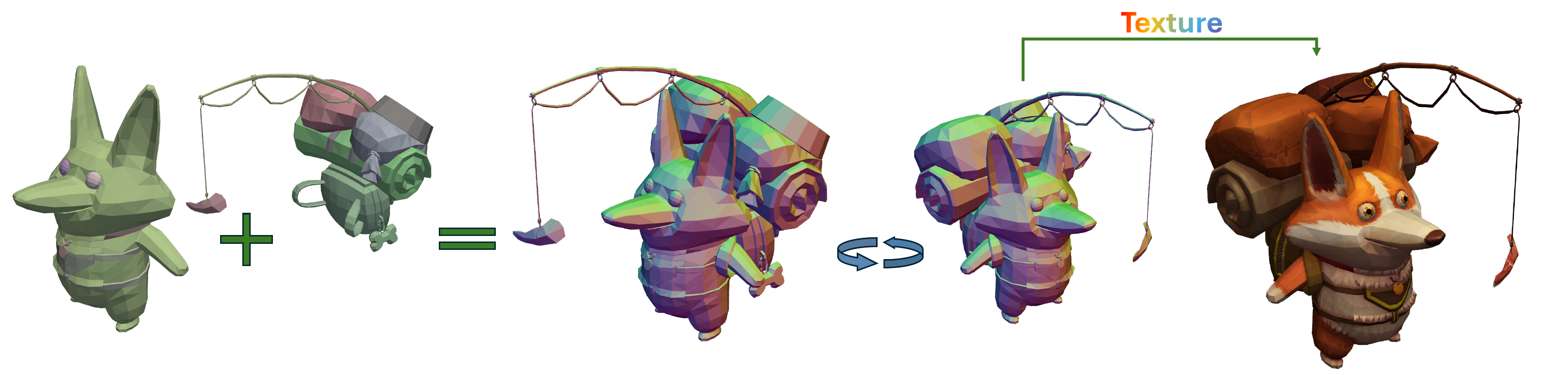}
  \vspace{-2mm}
  \caption{\textbf{Demonstration of FCT Editing.} \textit{\textbf{Assembly} of two geometric components, subsequent \textbf{Manipulation} (transformation/posing) of the combined object, and the \textbf{texture} can be recovered by voxel-wise RGB features attached on FCT.}
  }
  \label{fig:FCT_operations}
\end{figure}

The proposed representation near-losslessly converts meshes into voxel structures by directly fitting and storing the anchor points and connections. All operations that work for voxels can be easily applied to the FCT representation; meanwhile, additional features can be easily attached to each active voxel to represent features beyond geometry, such as partitioning, texture, materials, and semantic information. Examples of various editing operations are shown in Fig.~\ref{fig:FCT_operations}.

\vspace{-2mm}
\vspace{-2mm}

\vspace{-2mm}
\paragraph{Filtering}
Some deep learning tasks rely solely on surface information and require filtering out all hidden voxels. Ray-casting \cite{amanatides1987fast} of active voxels can easily determine visibility probabilities. This allows for the precise removal of voxels falling below a visibility threshold and, crucially, the corresponding deletion of their associated tokens from the FCT data. Similar voxel-wise filters, such as those based on quality or density thresholding, can be conducted using the same volumetric processing paradigm.

\vspace{-2mm}
\paragraph{Texture}
Due to the anchor in FCT tending to locate on the raw mesh, the closest-triangle search and linear projection can provide the corresponding UV coordinates for each. Since the FCT is a token-based representation, texture properties are handled efficiently by requiring only a few \textbf{additional channels attached} to each active voxel. Any attribute can therefore be sampled from the raw mesh and assigned back onto the active voxels via additional channels in the FCT.

\vspace{-2mm}
\paragraph{Manipulation}
Rotation and non-linear coordinate deformation, which are typically challenging for mesh-based representations, can be efficiently performed by applying the transformation matrix directly to the stored \textbf{anchor points} and recalculating the \textbf{connection} vectors or by leveraging the underlying voxel grid for {spatial indexing} and {transformation lookups}. 

\vspace{-3mm}
\paragraph{Partition and Assembly}
Geometric editing in the FCT is primarily achieved through a token-based Partition and Assembly mechanism. \textbf{Assembly} (component merging) is readily achieved by gathering the anchors of the constituent meshes on overlapped voxels. The process uses the \textbf{mean gather} of anchor positions and the \textbf{maximum gather} of semi-axis orientation to establish a coherent boundary and determine the properties of the resulting single component. Conversely, \textbf{Partition} (component separation) is realized by applying a geometric mask or semantic labeling to duplicate and segment the active anchors and connections into discrete, manageable token groups, enabling their separation and non-destructive editing.

\subsection{Faithful Contouring VAE}
To validate the effectiveness of applying Faithful Contouring as a high-fidelity representation for 3D modeling and generation, we employ a variational autoencoder (VAE~\cite{kingma2013auto}) to compress the Faithful Contouring representation. The VAE architecture draws inspiration from previous works~\cite{xiang2025structured, he2025triposf, sparc3d2025}, with modifications tailored for contour-preserving voxel representations.
\paragraph{Architecture.}
The encoder is composed of cascaded sparse 3D convolutional residual blocks, followed by lightweight local attention layers~\cite{wu2024point, sparc3d2025}, progressively compressing the input into a compact latent embedding. The decoder mirrors this structure, hierarchically upsampling the latent code and predicting reconstructed FCT.
To show the universality of such representation, we design a dual-mode input: the input can be features of either FCTs or points sampled from raw meshes.
In the \emph{auto-compression mode}, the architecture directly encodes FCTs into sparse latent codes and decodes them back to the original representation, enabling near-lossless compression.  
In the \emph{point-to-FCT mode}, we add a local attention layers before encoder blocks, to aggregate features of sampled points into corresponding voxels. These voxel-level features are subsequently encoded into latent codes to represent FCTs.
Such a design not only allows the network to preserve details of raw meshes and faithful contouring representations, but is able to reconstruct a structured voxelized contour representation directly from unstructured point sets, bridging modality gaps without explicit re-meshing operations. 


\paragraph{Training and Losses.}  
For the VAE training, we supervise the FCT features and corresponding occupancy logits for pruning redundant voxels during upsampling following~\cite{he2025triposf, sparc3d2025}. Specifically, we train the VAE with the following losses: Mean-Square-Error loss $\mathcal{L}_{\mathbf{x}}$ for the positions of anchors (e.g. relative offsets w.r.t.\ each voxel center), cosine similarity $\mathcal{L}_{\mathbf{n}}$ for their normals, binary cross entropy $\mathcal{L}_{\mathrm{axis}}, \mathcal{L}_{\mathrm{mask}}, \mathcal{L}_{\mathrm{occ}}$ for semi-axis codes, dual masks and occupancy of upsampled voxels, and KL divergence $\mathcal{L}_{\mathrm{KL}}$ for latent regularization. 
  
The final objective is a weighted combination:
\[
\mathcal{L} = \lambda_{\mathbf{x}}\mathcal{L}_{\mathbf{x}} 
+ \lambda_{\mathbf{n}}\mathcal{L}_{\mathbf{n}} 
+ \lambda_{\mathrm{axis}}\mathcal{L}_{\mathrm{axis}} 
+ \lambda_{\mathrm{mask}}\mathcal{L}_{\mathrm{mask}} 
+ \lambda_{\mathrm{occ}}\mathcal{L}_{\mathrm{occ}} 
+ \lambda_{\mathrm{KL}}\mathcal{L}_{\mathrm{KL}}.
\]

\section{Experiments}
\label{sec:experiments}

We evaluate \name{} on challenging real-world meshes and compare it against representative voxel-based and implicit reconstruction approaches. Our experiments focus on two aspects: (i) Representation fidelity (Sec.~\ref{exp:represent}) and (ii) VAE reconstruction quality (Sec.~\ref{exp:vae_quality}).

\subsection{Implementation Details}

\vspace{-2mm}
\paragraph{Implementation.} All core operators for the encoding (Alg.~\ref{alg:fct-encode}) and decoding (Alg.~\ref{alg:fct-decode}) stages of the proposed FCT representation are implemented as custom PyTorch and CUDA kernels to guarantee scalability and computational efficiency. Fitting and remeshing with resolutions under $1024^3$ run on a single RTX~3090 (24,GB), while fitting at $2048^3$ is completed on an RTX~A6000 (48,GB).

For the VAE baseline, we follow~\cite{sparc3d2025} and compress the FCT into an $8\times$ downsampled latent code for a fair comparison. The VAE is trained for 200K iterations on a cluster of 32 NVIDIA A100 GPUs. 

\vspace{-2mm}
\paragraph{Datasets.} For comparing the representation fidelity, we curate a rigorous benchmark by selecting complex, difficult cases from ABO~\cite{collins2022abo} and Objaverse~\cite{objaverse_xl} that specifically feature occluded parts, intricate geometries, and open surfaces. To further assess generalization capabilities, we also include diverse in-the-wild meshes with multiple disconnected components collected from independent online sources outside both major datasets. All input meshes are preprocessed by being normalized into the $[-1, 1]^3$ coordinate space with a $0.025$ margin.
We follow~\cite{xiang2025structured} to use about 400K data from Objaverse-XL~\cite{objaverse_xl} as the training data. For the VAE reconstruction comparisons, we further utilize Dora benchmark~\cite{Chen_2025_Dora} and Toys4k dataset~\cite{stojanov2021using} following~\cite{Chen_2025_Dora, xiang2025structured, he2025triposf}.

\paragraph{Baselines.} For the evaluation for capability of Faithful Contouring, we compare our approach against three representative families of 3D reconstruction methods: (i) UDF-based watertight surface reconstruction, commonly adopted in prior work~\cite{Chen_2025_Dora}; (ii) Flood-fill based Signed Distance Field (SDF) reconstruction; and (iii) deformed sparse-voxel SDFs introduced in SparC~\cite{sparc3d2025}.
We compare our VAE performance with Craftsman~\cite{li2024craftsman3d}, Dora~\cite{Chen_2025_Dora}, Trellis~\cite{xiang2025structured}, XCube~\cite{ren2024xcube}, SparseFlex~\cite{he2025triposf} and SparC~\cite{sparc3d2025}.
All baseline models are either implemented from scratch or adapted from publicly available code to ensure a fair comparison. The reported reconstruction metrics are computed directly from the outputs generated by each method, without applying any post-processing.

\subsection{Comparisons of Representations}
\label{exp:represent}

\vspace{-2mm}
\begin{figure}[!ht]
\centering
\includegraphics[width=\linewidth]{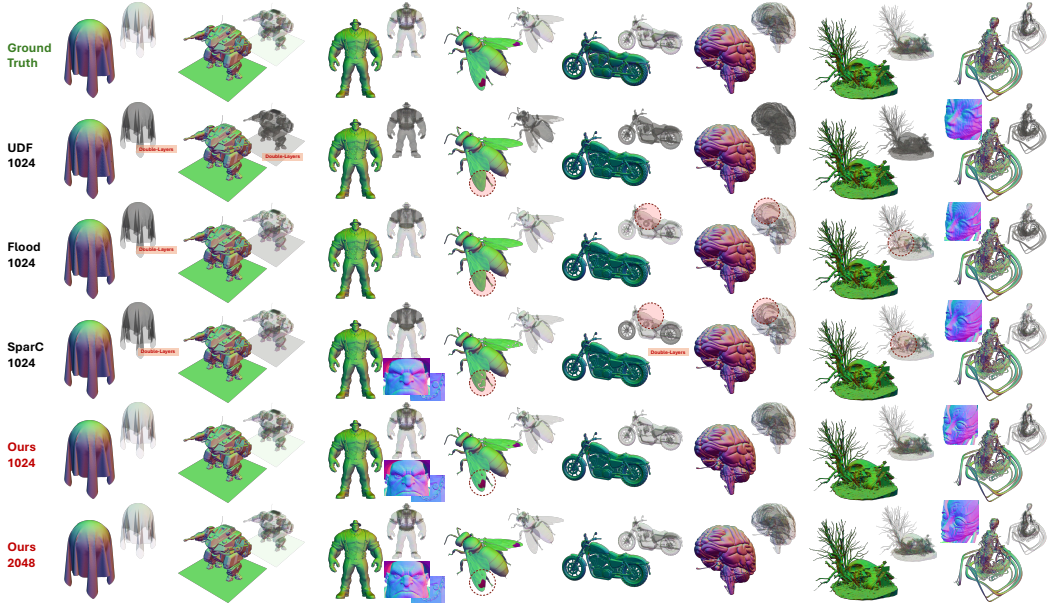}
\vspace{-2mm}
\caption{\textbf{Comparison of representations}. \textit{From top to bottom: Ground Truth, UDF (1024), Flood-Fill (1024), SparC~\cite{sparc3d2025} (1024), and our method (\name) at 1024 and 2048. Competing methods often suffer from \textbf{double-layer artifacts}, \textbf{loss of internal structures}, or \textbf{surface thickening} (red circles) and \textbf{voxel-lattice artifacts or bumping} on reconstructed faces. In contrast, \name{} generates clean, high-fidelity surfaces; represents open surfaces with \textbf{single-layers}; and faithfully preserves \textbf{fine details} and \textbf{internal geometries} across diverse categories, with higher resolution further improving details.}}
\label{fig:compare}
\end{figure}
\vspace{-2mm}

Reconstruction results from compared methods are shown in Figure~\ref{fig:compare}. \name{} produces clean surfaces with sharp features and faithfully retains internal structures. In contrast, UDF often yields low-fidelity results and produces characteristic \textbf{double-layer artifacts}, and Flood-fill often leads to undesirable \textbf{surface thickening} and a \textbf{loss of internal structures}. Furthermore, SparC, the current SOTA, despite using differentiable optimization on voxel corners, demonstrably struggles to reconstruct open geometries accurately and faithfully represent high-detail features such as the human face. More generally, all existing voxelized methods, including UDF, Flood-fill, and SparC, suffer from the inherent \textbf{grid bumping} artifacts introduced during the Marching Cubes remeshing, a limitation that \name{} uniquely circumvents via solving local QEF.

\vspace{-2mm}
\begin{table}[!ht]
\centering
\caption{\textbf{Quantitative comparison of different voxel representation}. 
All HD values are scaled by $10^{-2}$, and CD values by $10^{-4}$. All results are shown with $mean \pm std$.}
\vspace{-2mm}
\resizebox{\textwidth}{!}{%
\begin{tabular}{lcccccc}
\toprule
\textbf{Method} &
\textbf{HD $\downarrow$} &
\textbf{CD$_{P\to G}$ $\downarrow$} &
\textbf{CD$_{G\to P}$ $\downarrow$} &
\textbf{F1$_{0.01}$ $\uparrow$} &
\textbf{NCD $\downarrow$} &
\textbf{ANC $\uparrow$} \\
\midrule
UDF 512   & 0.12 $\pm$ 0.01 & 3.74 $\pm$ 0.22 & 0.65 $\pm$ 0.07 & 97.26 $\pm$ 1.26 & 0.89 $\pm$ 0.12 & 0.96 $\pm$ 0.03 \\
UDF 1024  & 0.20 $\pm$ 0.52 & 1.61 $\pm$ 0.02 & 0.42 $\pm$ 0.02 & 99.15 $\pm$ 0.07 & 0.88 $\pm$ 0.14 & 0.98 $\pm$ 0.04 \\
Flood 512   & 2.68 $\pm$ 11.32 & 4.71 $\pm$ 2.52 & 5.01 $\pm$ 20.38 & 95.69 $\pm$ 0.86 & 0.42 $\pm$ 0.40 & 0.95 $\pm$ 0.01 \\

Flood 1024  & 0.75 $\pm$ 3.53 & 1.68 $\pm$ 0.28 & 1.16 $\pm$ 5.38 & 98.85 $\pm$ 0.82 & 0.80 $\pm$ 0.21 & \textcolor{Burgundy}{\textbf{0.99 $\pm$ 0.01}} \\

SparC 512 & 2.57 $\pm$ 11.28 & 0.35 $\pm$ 0.28 & 4.48 $\pm$ 19.73 & 97.14 $\pm$ 1.01 & 0.44 $\pm$ 0.42 & 0.97 $\pm$ 0.02 \\

SparC 1024& 0.71 $\pm$ 1.26 & \textbf{0.30 $\pm$ 0.01} & 1.19 $\pm$ 3.24 & 98.50 $\pm$ 0.01 & 0.46 $\pm$ 0.35 & \textbf{0.98 $\pm$ 0.02} \\

\midrule
\rowcolor{pink!10}
\textbf{Ours 512} & 0.88 $\pm$ 0.12 & 0.32 $\pm$ 0.02 & 0.02 $\pm$ 0.01 & 99.15 $\pm$ 0.18 & \textbf{0.15 $\pm$ 0.14} & 0.93 $\pm$ 0.05 \\
\rowcolor{green!10}

\textbf{Ours 1024} & \textbf{0.11 $\pm$ 0.27} & 0.30 $\pm$ 0.02 & \textbf{0.01 $\pm$ 0.01} & \textbf{99.71 $\pm$ 0.08} & \textcolor{Burgundy}{\textbf{0.13 $\pm$ 0.13}} & 0.96 $\pm$ 0.03 \\

\rowcolor{blue!10}
\textbf{Ours 2048} &
\textcolor{Burgundy}{\textbf{0.11 $\pm$ 0.18}} &
\textcolor{Burgundy}{\textbf{0.24 $\pm$ 0.01}} &
\textcolor{Burgundy}{\textbf{$<$0.01}} &
\textcolor{Burgundy}{\textbf{99.99 $\pm$ 0.00}} &
{0.24 $\pm$ 0.16} &
0.97 $\pm$ 0.02 \\
\bottomrule
\end{tabular}}
\vspace{-2mm}
\label{tab:quantitative}
\end{table}

Quantitative evaluations corroborating these observations are summarized in Table~\ref{tab:quantitative}. 
We adopt standard surface metrics, where the bi-directional Chamfer Distance (CD) is decomposed into two complementary components: 
$\text{CD}_{P\to G}$ reflects \textit{completeness} by penalizing redundant or over-populated surface predictions, whereas 
$\text{CD}_{G\to P}$ measures \textit{accuracy} by quantifying the extent to which fine geometric structures in $G$ are successfully recovered by $P$.
This separation allows a more precise characterization of the representation’s ability to capture high-frequency details and internal cavities.

At a resolution of 1024, \name{} achieves the lowest Hausdorff Distance (${0.11 \pm 0.27 \times 10^{-2}}$) among all competitors, together with the minimum $\text{CD}_{G\to P}$ (${0.01 \pm 0.01 \times 10^{-4}}$), 
indicating accurate recovery of thin, sharp, and occluded structures that UDF, Flood-fill, and SparC frequently fail to reconstruct. 
Furthermore, the extremely high $\text{F}_{0.01}$ score (${99.71 \pm 0.08}$) confirms a robust precision–recall balance and demonstrates that \name{} does not suffer from the surface thickening or volume swelling commonly introduced by implicit- or fill-based schemes.

At a resolution of 2048---a scale unattainable by prior voxel representations due to global optimization requirements or memory constraints---\name{} further improves all error terms and establishes a new state of the art:
$\text{HD} = {0.11 \pm 0.18 \times 10^{-2}}$,
$\text{CD}_{G\to P} < {0.01 \times 10^{-4}}$,
$\text{F}_{0.01} = {99.99 \pm 0.002}$,
$\text{NCD} = {0.24 \pm 0.16}$,
and $\text{ANC} = {0.97 \pm 0.02}$. 
Such exceptionally low $\text{CD}_{G\to P}$ values quantitatively verify that \name{} faithfully retains internal cavities and delicate high-curvature features, rather than implicitly smoothing or filling them as in UDF and Flood-fill. 
Notably, \name{} is the \textbf{only} voxel-based method capable of scalable reconstruction at $2048^3$ and achieves $<10^{-4}$ bi-directional CD relative to all baselines, conclusively demonstrating superior geometric fidelity and unparalleled scalability.

\subsection{Comparison of Reconstructions}
\label{exp:vae_quality}

\begin{figure}[!ht]
\centering
\includegraphics[width=\linewidth]{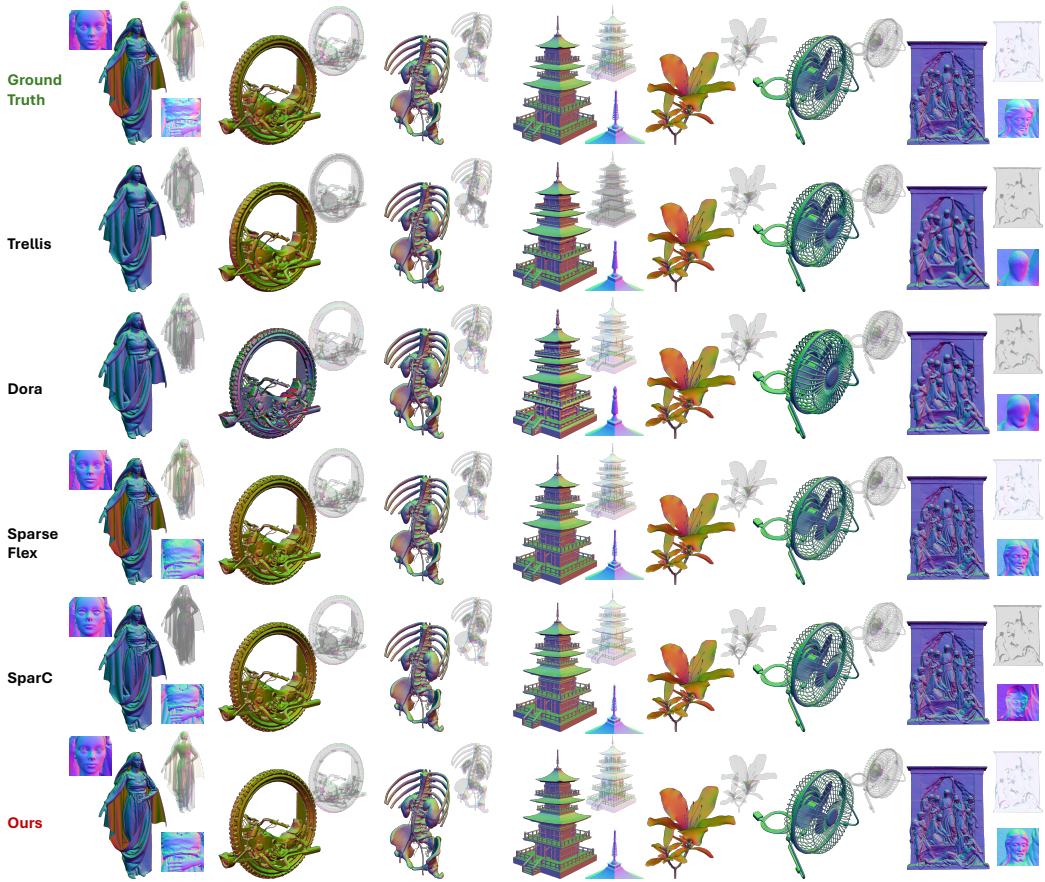}
\vspace{-2mm}
\caption{\textbf{Comparison of VAE reconstructions}.\textit{ Our method demonstrates
superior performance in reconstructing complex shapes, open surfaces, and interior structures, compared to existing VAEs.
}}
\label{fig:vae}
\end{figure}

Quantitative and qualitative results of VAE reconstruction are presented in ~\ref{tab:vae} and ~\ref{fig:vae}, respectively. Following prior work~\cite{he2025triposf}, we evaluate VAE performance using Chamfer Distance (CD) and F-score with thresholds of 0.1 and 0.001. The reported values are scaled by $10^4$ and $10^2$, respectively. For a fair comparison, we evaluate all methods not only on the entire datasets, but also on a subset of watertight meshes, as Dora~\cite{Chen_2025_Dora}, Craftsman~\cite{li2024craftsman3d}, and Sparc3D~\cite{sparc3d2025} require pre-processed watertight inputs and perform poorly on non-watertight geometries. As shown in Table~\ref{tab:vae}, our method, FaithC, significantly outperforms recent state-of-the-art approaches such as SparseFlex and Sparc3D. Specifically, FaithC achieves approximately 93\% lower CD and a 35\% improvement in F-scores. Visual comparisons in Figure~\ref{fig:vae} further illustrate that FaithC effectively preserves thin structures, sharp edges, and complex intersections between mesh components, whereas previous SDF-based methods are often unable to express these details, an inherent limitation of their underlying formulation.

\vspace{-2mm}
\begin{table}[!ht]
\centering
\caption{\textbf{Quantitative results of VAE reconstruction quality.}
The ``/'' separates results over the full dataset vs.\ the watertight subset.
\textsuperscript{\dag} indicates our re-implementation. We specify the compression schemes for different VAEs, where ``Vec.'' indicates compression using vecset~\cite{zhang20233dshape2vecset}, and ``Vox. $N\times$'' indicates compression to downsampled voxels with $1/{N}$-lower resolution.}
\vspace{-2mm}

\footnotesize
\setlength\tabcolsep{2pt}
\resizebox{\linewidth}{!}{%
\begin{tabular}{l|c|ccc|ccc}
\toprule
\multirow{2}{*}{\textbf{Method}} &
\multirow{2}{*}{\begin{tabular}{c}\textbf{Comp.}\\\textbf{Scheme}\end{tabular}} &
\multicolumn{3}{c|}{\textbf{Toys4K}} &
\multicolumn{3}{c}{\textbf{Dora Benchmark}} \\
\cmidrule(lr){3-5} \cmidrule(lr){6-8}
& & $\mathbf{CD}\!\downarrow$ & $\mathbf{F1}_{\mathbf{0.001}}\!\uparrow$ & $\mathbf{F1}_{\mathbf{0.01}}\!\uparrow$
  & $\mathbf{CD}\!\downarrow$ & $\mathbf{F1}_{\mathbf{0.001}}\!\uparrow$ & $\mathbf{F1}_{\mathbf{0.01}}\!\uparrow$ \\
\midrule
Craftsman~\cite{li2024craftsman3d}     & Vec.   & 13.08/4.63 & 10.13/15.15 & 56.51/85.02 & 13.54/2.06 & 6.30/11.14 & 73.71/91.95 \\
Dora~\cite{Chen_2025_Dora}            & Vec.   & 11.15/2.13 & 17.29/26.55 & 81.54/93.84 & 16.61/1.08 & 13.65/25.78 & 78.73/96.40 \\
Trellis~\cite{xiang2025structured}    & Vox. 4$\times$   & 12.90/11.89 & 4.05/4.93 & 59.65/64.05 & 17.42/9.83 & 3.81/6.20 & 62.70/71.95 \\
XCube~\cite{ren2024xcube}             & Vox. 4$\times$   & 4.35/3.14 & 1.61/13.49 & 74.65/79.62 & 4.74/2.37 & 1.31/0.84 & 75.64/86.50 \\
3PSDF$^{\dag}$~\cite{chen20223psdf}   & Vec.   & 4.51/3.69 & 11.33/14.10 & 81.70/86.13 & 7.45/1.68 & 7.52/12.50 & 79.43/91.17 \\
SparseFlex$_{1024}$~\cite{he2025triposf} & Vox. 4$\times$   & 1.33/0.60 & 25.95/\textbf{35.69} & 92.30/96.22 & 0.86/0.12 & 25.71/\textcolor{Burgundy}{\textbf{39.50}} & 94.71/99.14 \\
SparC$_{1024}^{\dag}$~\cite{sparc3d2025} & Vox. 8$\times$ & 11.42/9.80 & 18.26/26.44 & 74.72/83.67 & 2.67/0.97 & 25.64/33.34 & 94.95/97.55 \\
\midrule
\rowcolor{pink!10}
\textbf{Ours$^{~pc}_{~512}$}    & Vox. 8$\times$ & 0.59/0.20 & 28.72/33.20 & 97.06/98.98 & 0.09/0.06 & 27.01/33.76 & 99.76/99.93 \\
\rowcolor{green!10}
\textbf{Ours$_{~512}$ }        & Vox. 8$\times$ & \textbf{0.57}/\textbf{0.18} & \textbf{30.33}/34.60 & \textbf{97.15}/\textbf{99.09} & \textbf{0.07}/\textcolor{Burgundy}{\textbf{0.05}} & \textbf{28.75}/34.85 & \textbf{99.88}/\textcolor{Burgundy}{\textbf{99.99}} \\
\rowcolor{blue!10}
\textbf{Ours$_{~1024}$}        & Vox. 8$\times$ & \textcolor{Burgundy}{\textbf{0.46}}/\textcolor{Burgundy}{\textbf{0.13}} & \textcolor{Burgundy}{\textbf{34.91}}/\textcolor{Burgundy}{\textbf{38.45}} & \textcolor{Burgundy}{\textbf{97.89}}/\textcolor{Burgundy}{\textbf{99.39}} &
\textcolor{Burgundy}{\textbf{0.06}}/\textcolor{Burgundy}{\textbf{0.05}} & \textcolor{Burgundy}{\textbf{30.80}}/\textbf{36.03} & \textcolor{Burgundy}{\textbf{99.97}}/\textcolor{Burgundy}{\textbf{99.99}} \\
\bottomrule
\end{tabular}
}
\vspace{-1mm}
\label{tab:vae}
\end{table}

\subsubsection{Effects of Different Resolutions and Input Features}

We further investigate reconstruction performance under different input modalities—namely point clouds and FCT features. Although VAE-based conversion from point clouds to FCT features offers broader applicability, results in Table~\ref{tab:vae} indicate that at a resolution of 512, direct compression of FCT features yields superior reconstruction quality compared to point cloud compression. We attribute this to the limited expressiveness of sparse point clouds, which restricts the amount of structural information available for reconstruction. In contrast, FCT features preserve full geometric information, leading to more accurate reconstructions.

Notably, even at the lower resolution of 512, FaithC substantially outperforms state-of-the-art methods such as SparseFlex and Sparc3D at resolution 1024, demonstrating the high capacity and fidelity of our approach when reconstructing diverse kinds of meshes.
As the resolution increases to 1024, the reconstruction performance of FaithC VAE can be even further improved.

\vspace{-2mm}

\section{Conclusion}

We introduced \textbf{Faithful Contouring}, a near-lossless and remeshable voxel representation that directly encodes meshes into sparse contour tokens without distance fields or iso-surface extraction. This design preserves sharp geometry, open surfaces, and internal cavities, and scales beyond $2048^3$ via fully local parallel computation. 
Combined with a dual-mode autoencoder, \name{} supports detail-preserving reconstruction from contour tokens or raw point sets, and consistently surpasses existing implicit and sparse-voxel methods in accuracy and efficiency. To the best of our knowledge, this is the first voxel representation that removes the dependency on both SDF conversion and Marching Cubes, enabling faithful reconstruction.

\paragraph{Limitations.}
Although \name{} achieves high fidelity, complex cases such as severe self-intersections or multiple closely spaced thin layers can introduce ambiguous anchors, leading to small local drifts. 
Moreover, the VAE does not yet fully exploit the expressive capacity of FCT, particularly for highly irregular structures. The smoothness and sharpness of decoded FCT are slightly decreased compared to orginal fitting. 

\paragraph{Future Work.}
Future work will focus on improving robustness in these challenging geometric scenarios, developing differentiable contouring and rendering to integrate with gradient-based learning, and exploring dynamic resolution to better allocate capacity around thin structures. 
In addition, we aim to leverage contour tokens as a structured latent representation for high-precision 3D generation, extending \name{} beyond reconstruction toward scalable generative modeling.

\bibliography{iclr2025_conference}
\bibliographystyle{iclr2025_conference}

\end{document}